\documentclass[journal]{IEEEtran}
\makeatletter
\def\endthebibliography{%
	\def\@noitemerr{\@latex@warning{Empty `thebibliography' environment}}%
	\endlist
}
\makeatother
\usepackage{amsmath,graphicx}
\usepackage[caption=false,font=footnotesize]{subfig}
\usepackage{amsthm}
\usepackage{array}
\usepackage{float}
\newcolumntype{P}[1]{>{\centering\arraybackslash}p{#1}}
\newcolumntype{M}[1]{>{\centering\arraybackslash}m{#1}}

\usepackage{cite}
\usepackage{amsmath,amssymb,amsfonts}
\usepackage{bm}
\usepackage{graphicx}
\usepackage{textcomp}
\usepackage{algorithm}
\usepackage{algpseudocode}
\usepackage{multicol}

\def\BibTeX{{\rm B\kern-.05em{\sc i\kern-.025em b}\kern-.08em
		T\kern-.1667em\lower.7ex\hbox{E}\kern-.125emX}}

\ifCLASSINFOpdf
\else
\fi
\begin{document}
%
\title{Latent Code-Based Fusion: A Volterra Neural Network Approach}

%
%
%

\author{Sally~Ghanem, Siddharth~Roheda, and~Hamid~Krim\\
ECE Department, NCSU, Raleigh, NC, 27695 \IEEEmembership{}
}
\maketitle

\begin{abstract}
We propose a deep structure encoder using the recently introduced Volterra Neural Networks (VNNs) to seek a latent representation of multi-modal data whose features are jointly captured by a union of subspaces. The so-called self-representation embedding of the latent codes leads to a simplified fusion which is driven by a similarly constructed decoding. The Volterra Filter architecture achieved reduction in parameter complexity is primarily due to controlled non-linearities being introduced by the higher-order convolutions in contrast to generalized activation functions. Experimental results on two different datasets have shown a significant improvement in the clustering performance for VNNs auto-encoder over conventional Convolutional Neural Networks (CNNs) auto-encoder. In addition, we also show that the proposed approach demonstrates a much-improved sample complexity over CNN-based auto-encoder with a superb robust classification performance. 
\end{abstract}

\begin{IEEEkeywords}
Sparse learning, Computer vision, Information fusion, Subspace clustering.
\end{IEEEkeywords}

%
\IEEEpeerreviewmaketitle

\section{Introduction}
 \label{sec:intro}
 
 On account of recent advances in sensor technology, multimodal data has become widely available and useful. Additional modalities can often provide supplementary information about the targets/classes of interest. A principled integration of multi-modal sensor data is expected to improve the quality of extracted features. Multi-modal fusion has been extensively used in different applications including but not limited to image fusion \cite{hellwich2000object}, target recognition \cite{korona1996model}\cite{ghanem2018information} \cite{ghanem2020robust}, speaker recognition \cite{soong1988use} and handwriting analysis \cite{xu1992methods}. In addition, CNNs have been extensively utilized for multimodal data analysis as in \cite{ngiam2011multimodal}, \cite{roheda2018cross},\cite{roheda2020robust}, and \cite{roheda2018decision}. However, the complexity of implementing a multi-modal fusion network still persists.\\ 
 Convolutional Neural Networks (CNNs) have widely been adopted in deep learning for analyzing visual images in many applications. These application include but are not limited to image processing, segmentation, and segmentation. However, the complexity and cost of implementing CNNs can be limiting in many applications. The efficient and recently proposed Volterra Neural Networks (VNNs) \cite{roheda2020conquering} was aimed to primarily address these limitations and also overcome the CNN over parametrization problem. To control the non-linearities intentionally induced in the network, a judicious setting of the degree of interactions between the delayed input samples of the data is induced. The cascaded implementation proposed in \cite{roheda2020conquering} has shown to significantly reduce the number of parameters needed for training the network in comparison to conventional neural networks. In addition to reducing the network complexity, Volterra Neural Networks (VNNs) have a more tractable and comprehensible structure. This represents a significant departure from the work in  \cite{zoumpourlis2017non} and \cite{kumar2011trainable}, as the strategy of convolution together with an understanding of the inherent complexity of a naive approach resulted in the success of this new outlook. These prior approaches had limited the degree of the non-linearities up to certain value to avoid the explosive complexity, but thereby defeating the initial goal of overcoming the limitations of existing CNNs. The cascaded implementation in \cite{roheda2020conquering} has been shown to alleviate this limitation through applying the second order filter repeatedly applied until the desired order is attained. \\
 Inspired by the success of VNNs in deep learning \cite{roheda2020conquering}, we propose an efficient implementation of the Deep Multi-modal Subspace clustering\cite{abavisani2018deep}, using Volterra filters. More specifically, the CNNs are replaced with VNNs which controls the introduced non-linearities via high order convolutions instead of using highly non-linear activation functions as in \cite{abavisani2018deep} architecture. Moreover, we propose additional techniques to significantly reduce the complexity of a VNN inspired Multi-Modal Subspace Clustering Auto-Encoder as compared with \cite{abavisani2018deep} to a fraction of the number of parameters used by CNNs while retaining a comparable clustering performance.\\
 The rest of the paper is organized as follows, in Section 2, we provide an overview for DMSC \cite{abavisani2018deep}, which is based on CNNs. In section 3, we provide the problem formulation along with the our proposed approach, namely, Volterra Multi-Modal Subspace Clustering Auto Encoder (VMSC-AE). In Section 4, we present a substantiative validation along with experimental results of our approach, while Section 5 provides concluding remarks.
\section{Affinity Fusion Deep Multimodal Subspace Clustering}

In this Section, we provide a brief overview of the Affinity Fusion Deep Multimodal Subspace Clustering (AFDMSC) which was proposed in \cite{abavisani2018deep}. AFDMSC network is composed of three main parts: a multimodal encoder, a self-expressive layer, and a multimodal decoder. The output of the encoder contributes to a common latent space for all modalities. AFDMSC utilizes the self-expressive property, which was presented in \cite{elhamifar2013sparse} and \cite{bian2015bi}. This property ensures acquiring the latent space structure that reveals the relationships between data points in each cluster. The self-representation property entails the representation of each sample as a linear combination of all other samples from the same subspace/cluster. The self-expressive property is enforced through a fully connected layer between the encoder and the decoder results in one common set of weights for all the data sensing modalities. The reconstruction of the input data by the decoder, yields the following loss function to secure the proper training of the self-expressive network, 
\begin{equation}
\begin{split}
\min_{\mathbf{W}/w_{kk}=0} & \parallel \mathbf{W} \parallel_2+ \frac{\gamma}{2} \sum_{t=1}^{T} \parallel \mathbf{X}(t)-\mathbf{X_r}(t)\parallel_F^2+ \\
&\frac{\mu}{2} \sum_{t=1}^{T}\parallel \mathbf{L}(t)-\mathbf{L}(t)\mathbf{W}\parallel_F^2,
\end{split}
\end{equation}
where $\mathbf{W}$ represents the parameters of the self expressive layer, $\mathbf{X}(t)$ is the input to the encoder, $\mathbf{X_r}(t)$, denotes the output of the decoder and $\mathbf{L}(t)$ denotes the output of the encoder. $\mu$ and $\gamma$ are regularization parameters. An overview for the DMSC approach is illustrated in Figure 1.
\begin{figure}[!htb]
	\centering
	\includegraphics[width=9cm,height=4.5cm]{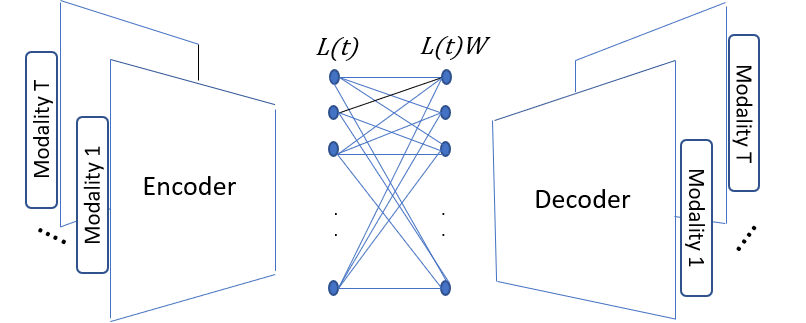}

	\caption{Deep Multimodal Subspace Clustering}.
\end{figure}

\section{Volterra Filter Subspace Clustering}

\subsection{Problem Formulation}
We consider a set of data observations indexed by \textit{k = 1, 2, ....., n} and assume \textit{T} data modalities, indexed by \textit{t = 1, 2, 3,....., T}. Each data modality $\mathbf{X}(t)=[\mathbf{x}_1(t) \  \mathbf{x}_2(t) \ ....... \ \mathbf{x}_n(t)] \in \mathbb{R}^{m \times n}$. The goal is to partition $\mathbf{X}(t)$ into clusters that can be well-represented by low-dimensional subspaces. Mathematically, this is equivalent to partitioning  $\mathbf{X}(t)$ as \{$\mathbf{X}^1(t),\mathbf{X}^2(t), ........,\mathbf{X}^P(t)$\} of [$n$], where $P$ is the number of clusters indexed by $p$, such that there exist linear subspaces $\mathbf{S}^p(t) \subset \mathbb{R}^m$ with dim($\mathbf{S}^p(t)$) $ \ll m$.

\subsection{Volterra Filter Auto-encoder}
In this section, we provide the fundamental concepts of our proposed approach which we refer to as Volterra Multimodal Subspace clustering auto-encoder (or VMSC). Our framework learns an efficient latent representation of multi-modal data whose features are jointly captured by a union of subspaces. Our approach was inspired by the Volterra series\cite{volterra1959theory}, which is a model for non-linear behavior similar to the Taylor series. Volterra series represents a functional expansion of non-linear and time-invariant systems. It differs from the Taylor series in its ability to capture memory. In other words, Taylor series can be used for approximating the response of a non-linear system  if the output depends strictly on the input at that particular time. In contrast to Volterra series, whose  output depends on the input to the system at all other times. Volterra filters (VF) are based on the Volterra series, VF describe a non-linear system via higher order convolutions. The relation between the output and input of the Volterra filter can be expressed as follows, 
\begin{equation}
\begin{split}
&\sum_{\tau_1=0}^{L-1} H^{1}_{\tau_1}x_{t-\tau_1}+\sum_{\tau_1, \tau_2=0}^{L-1} H^{2}_{\tau_1,\tau_2}x_{t-\tau_1} x_{t-\tau_2}+...\\
&+\sum_{\tau_1, \tau_2, ..., \tau_k=0}^{L-1} H^{k}_{\tau_1,\tau_2, ..., ,\tau_k} x_{t-\tau_1} x_{t-\tau_2}...x_{t-\tau_k}
\end{split}
\end{equation}
where $L$ is the number of terms in the filter memory (also referred to as the filter length), $H^k$ are the weights for the $k^{th}$ order term. The linear term in the previous equation is actually similar to a convolutional layer in CNNs. Generally speaking, non-linearities in CNNs are often introduced by activation functions, and not in the convolutional layer, while VNNs  introduce non-linearities in the convolutional layers. Volterra Neural Network (VNNs) has been recently proposed to control the non-linearities introduced in the network and to overcome the CNN over parametrization problem. The  VF architecture introduces controlled non-linearities in the Neural Networks in contrast to non-linear activation functions, which provide infinite non-linearity into the neural network which can not be easily truncated. Non-linear activation functions, such as Relu and Sigmoid, are often utilized in CNNs to act as a gate in between the input of a neuron and its output. However, in some cases, the unnecessary nonlinearities introduced by those activation functions in CNNs induce useless or irrelevant information to the network, which might confuse the classifier. In this work, we propose a multi-modal autoencoder using the recently introduced Volterra Neural Networks (VNNs) \cite{roheda2020conquering} to seek a latent representation of multi-modal data whose features are jointly captured by a union of subspaces. More specifically, we replace the CNNs in our network  with VNNs to control the introduced non-linearities and to non-linearly map the data points to a latent space that is well-adapted to subspace clustering in an unsupervised manner. The VMSC exploits the self-expressive property presented in \cite{elhamifar2013sparse} and \cite{bian2015bi} to acquire the latent space structure that reveals the relationships between data points in each cluster. In the following, we elaborate on the structure of the Volterra structure-based multi-modal auto-encoder. As noted earlier for (AFDMSC)\cite{abavisani2018deep}, the network has 3 main components, namely a multi-modal encoder, a self-expressive layer, and a multi-modal decoder. 
 
 The encoder in this work replaces the standard CNNs with the Volterra Neural Networks (VNNs) developed in  \cite{roheda2020conquering}. The multi-modal encoder consists of $T$ parallel Volterra NNs. Each branch of the encoder processes one of the modalities and extracts relevant features. The $T$ feature maps are subsequently concatenated, promoting the goal of obtaining a common latent space. The second component of the auto encoder is the self-expressive layer, the goal of which is to enforce the self-expressive property \cite{elhamifar2013sparse} among the concatenated features. This is enforced by a fully connected layer which operates on the concatenated output of the encoder. The last stage is the decoder which reconstructs the input data from the self-expressive layers' output. The objective functional sought through this approximation network is reflected in Equation (3).  A diagram showing our algorithm is depicted in Figure 2. 
\begin{equation}
\begin{split}
\min_{\mathbf{W}/w_{kk}=0} &\parallel \mathbf{W} \parallel_1+ \frac{\gamma}{2} \sum_{t=1}^{T} \parallel \mathbf{X}(t)-\mathbf{X_r}(t)\parallel_F^2+\\ 
&\frac{\mu}{2} \parallel \mathbf{L_{concat}}-\mathbf{L_{concat}} \mathbf{W}\parallel_F^2,
\end{split}
\end{equation}

where $\mathbf{X_r}(t)$ represents the reconstructed data corresponding to modality $t$, and $\mathbf{L_{concat}}$ is the concatenation of $\mathbf{L}(1), \mathbf{L}(2), ..., \mathbf{L}(T)$, and $\mathbf{L}(t)$ is the output of the encoder corresponding to modality $t$ . $\mathbf{W}$ is the sparse weight function that ties the concatenated features. The above cost function is optimized in Tensorflow using the adaptive momentum based gradient descent method (ADAM) \cite{kingma2014adam}. Under a suitable arrangement/permutation of the data realizations, the sparse coefficient matrix $\mathbf{W}$ is an $n \times n$ block-diagonal matrix with zero diagonals provided that each sample is represented by other samples only from the same subspace. More precisely, $\mathbf{W_{ij}}=0$ whenever the indices $i,j$ correspond to samples from different subspaces. As a result, the majority of the elements in $\mathbf{W}$ are equal to zero. $\| \|_1$ denotes the $l_1$ norm, i.e. the sum of absolute values of the argument and is used in order to ensure that the $\mathbf{W}$ matrix is sparse. 
 
To proceed with distinguishing the various classes in  an unsupervised manner, we first evaluate the affinity matrix as follows,
\begin{equation}
\mathbf{A}=\mathbf{W}+\mathbf{W}^T,
\end{equation}
where $\mathbf{A} \in \mathbb{R}^{n \times n} $. We subsequently use spectral clustering as detailed in \cite{ng2002spectral} to find the clusters underlying the multi-modal data.

\begin{figure*}[!htb]
	\begin{multicols}{1}
		\includegraphics[width=2\linewidth]{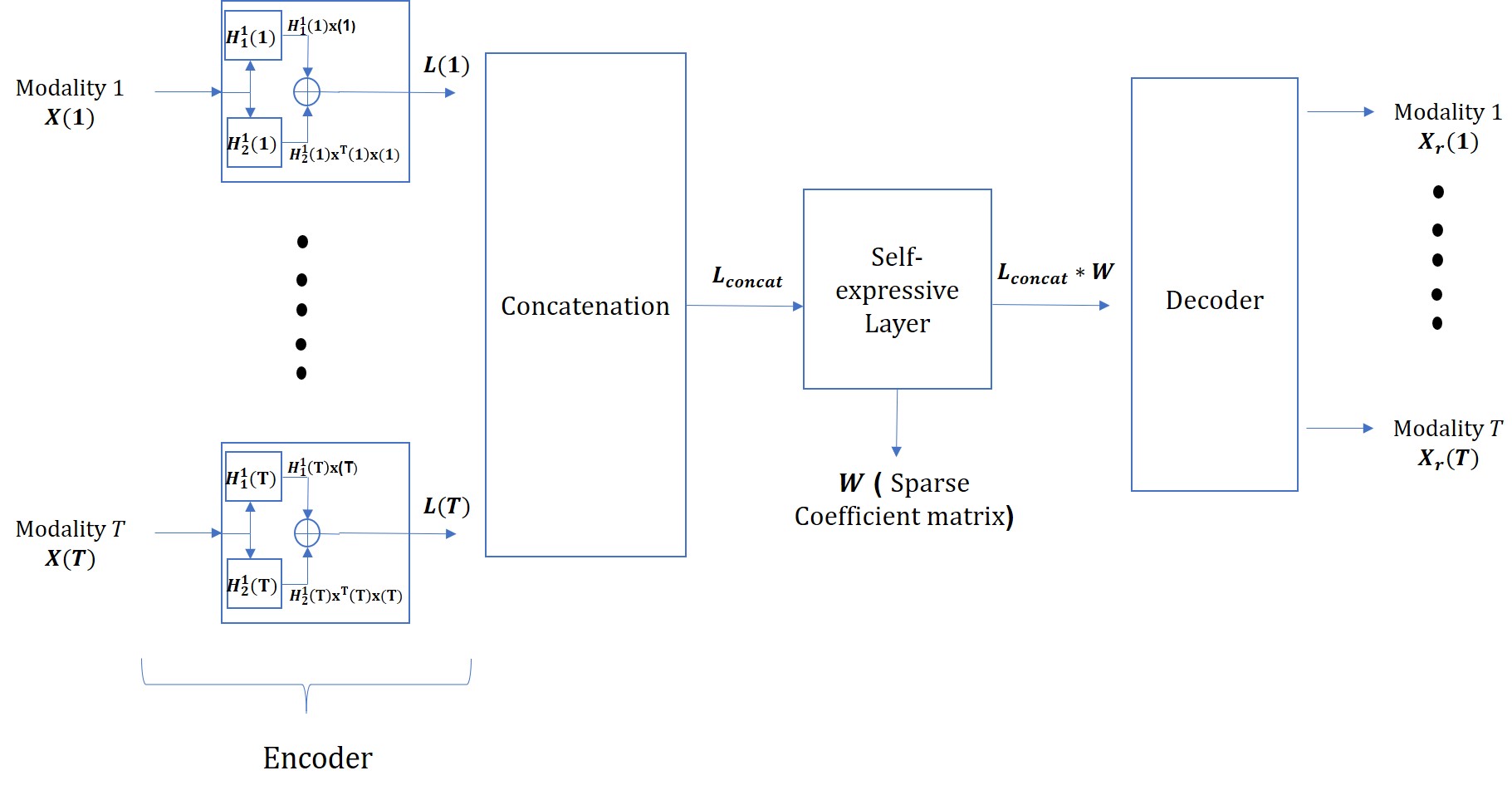}\par 
	\end{multicols}
	\caption{Volterra Neural Network Auto-encoder}
\end{figure*}

\subsection{Reducing the Model Complexity}

In this subsection, we will introduce two different solutions to further reduce the number of parameters required in training the auto-encoder network. The total number of parameters required for training the auto-encoder network is dominated by the self-expressive layer parameters, which is a function of $O(N^2)$, where $N$ is the number of samples in the dataset. As a result, this may lead to a longer training time and require a lot of computational resources. In section IV, we attempt to reduce the number of parameters needed by randomly removing a ratio of the edges in the self-expressive layer and train the network with the remaining edges. By eliminating the appropriate number of edges and setting them equal to zero, the clustering performance should not be highly affected. The reason is the fact that the self-expressive coefficient matrix, $\mathbf{W}$ is sparse and has a block-diagonal structure. Therefore, most of the edges will eventually be equal to zero.

 The second approach we evaluate in section IV is referred to as cyclic sparsely connected (CSC) layers \cite{hosseini2019complexity}. This approach was employed as an overlay for fully connected (FC) layers whose number of parameters, $O(N^2)$, can dominate the parameters of the entire deep neural network model. The CSC layers are composed of a few sequential layers, referred to as support layers, which result in full connectivity between the inputs and outputs of each CSC layer. The sparsely connected layers are composed of a sequence of $L$ layers: an Input layer, $L-1$ layers in between referred to as support layers, and an output layer. Assume having $N$ nodes and the fan-out of every node, as well as the fan-in of every layer, is equal to $F$. Therefore, the total number of edges, $E$, in the CSC layers structure is as follows,
\begin{equation}
E=NFL,
\end{equation}
Every layer in the CSC structure is defined by an adjacency matrix, $A_i$ s.t. $i=1,...,L$, whose length and width is equal to $N$, where $A(i, j)$ indicate the number of edges that connect the input node $i$ to the output node $j$. As a result, $NF$ out of $N^2$ elements of each adjacency matrix corresponding to one support layer are equal to 1 and the rest of the elements are zeros. The connectivity, $C$, is defined as the number of paths between any pair from the diagram input and output layers. Therefore,
\begin{equation}
F^L=NC
\end{equation}
From Eqns. (5) and (6), it can be deduced that the complexity decreased from $O(N^2)$ to $O(NlogN)$. The relationship between the number of edges and the number of nodes is as follows,
\begin{equation}
E=NFlog_F(NC)
\end{equation}

An adjacency matrix $A_i$ is defined for every support layer $i$ in the CSC layers we use to replace the fully connected layer. Every support layer has a generator polynomial $p_i(x)$, which is composed of $F$ terms to generate a cyclic adjacency matrix of block length $N$ as explained in \cite{hosseini2019complexity}. The generator polynomial corresponding to each support cyclic matrix constructs the first row and every next row of the matrix is a cyclic right shift of its previous row. In \cite{hosseini2019complexity}, two different factorization approaches were proposed for constructing the support layers. For our problem, we assume that the connectivity, $C$, is equal to 1. We follow the first approach proposed in \cite{hosseini2019complexity} to construct the generator polynomial $p_i(x)=\sum_{i=0}^{N-1}$ as follows,
\begin{equation}
\sum_{i=0}^{N-1}x^i=\prod_{i=0}^{L-1} p_i(x)~s.t.~    p_i(x)=\sum_{j=0}^{F-1} x^{S_i.j}, S_i=F^i,
\end{equation}
The CSC layers are described by the polynomial function by assigning each $p_i(x)$ to each support layer $i$. $S_i$ is the stride value which specifies the distance between elements of value 1s in the first row of the support matrix of layer $i$.

\section{EXPERIMENTAL RESULTS } 
\subsection{Dataset description}
To substantiate the discussed approach along with the various steps, we select two different datasets. The first one is the Extended Yale Dataset \cite{lee2005acquiring}. This dataset has been used extensively in subspace clustering as in  \cite{elhamifar2013sparse,liu2012robust}. The dataset is composed of 64 frontal images of 38 individuals under different illumination conditions. In this work, we utilize the augmented data used in \cite{abavisani2018deep}, where facial components such as left eye, right eye, nose and mouth have been cropped to represent four additional modalities. Images corresponding to each modality have been cropped to a size of 32$\times$32 A sample image for each modality is shown in Figure (3). 

\begin{figure}[!htb]
	\centering
	\subfloat[Face. ]{\includegraphics[width=3.5cm,height=3.1cm]{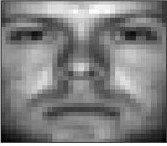}\label{fig:f1}}
	\hfill
	\subfloat[Left eye. ]{\includegraphics[width=3.5cm,height=3.1cm]{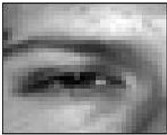}\label{fig:f2}}
	
	\centering
	\subfloat[Right eye. ]{\includegraphics[width=3.5cm,height=3.1cm]{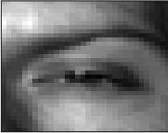}\label{fig:f3}}
	\hfill
	\subfloat[Mouth. ]{\includegraphics[width=3.5cm,height=3.1cm]{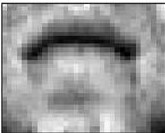}\label{fig:f4}}
	\centering
	\hfill
	\subfloat[Nose. ]{\includegraphics[width=3.5cm,height=3.1cm]{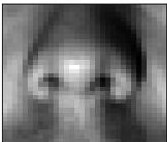}\label{fig:f5}}

	\caption{Sample images from the Augmented Extended Yale-B Dataset.}
\end{figure}

The second validation dataset we use is the ARL polarimetric face dataset \cite{hu2016polarimetric}, which is composed of facial images for 60 individuals in the visible domain and in four different polarimetric states. The dataset was collected using a polarimetric long-wave infrared imager, to facilitate cross-spectrum face recognition research. Different polarization states of thermal emissions provide additional geometric and textural facial details, which can be used to improve face identification. The Stokes parameters S0, S1, S2, and S3 are often used to represent polarization-state information. They are collected by measuring the radiant intensity transmitted through a polarizer that rotates at different angles. S0 represents the conventional total intensity thermal image, S1 captures the horizontal and vertical polarimetric information, and S2 captures the diagonal polarimetric information.  S1 and S2 capture orthogonal, yet complementary, polarimetric information. The degree-of-linear-polarization (DoLP) describes the portion of an electromagnetic wave that is linearly polarized. All the images are spatially aligned for each subject. We have also resized the images to 32$\times$32 pixels. Sample images from this dataset are shown in Figure (4).

\begin{figure}[!htb]
	\centering
	\subfloat[Visible. ]{\includegraphics[width=3.5cm,height=3.1cm]{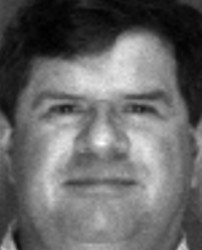}\label{fig:f6}}
	\hfill
	\subfloat[DoLP. ]{\includegraphics[width=3.5cm,height=3.1cm]{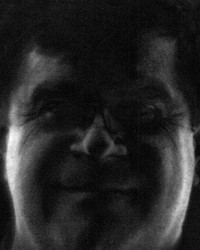}\label{fig:f7}}
	
	\centering
	\subfloat[S0. ]{\includegraphics[width=3.5cm,height=3.1cm]{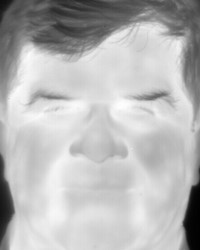}\label{fig:f8}}
	\hfill
	\subfloat[S1. ]{\includegraphics[width=3.5cm,height=3.1cm]{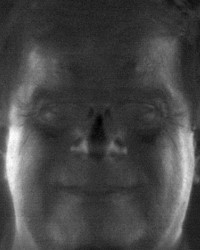}\label{fig:f9}}
	\centering
	\hfill
	\subfloat[S2. ]{\includegraphics[width=3.5cm,height=3.1cm]{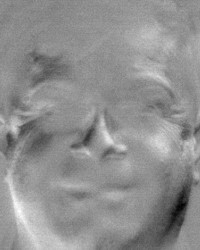}\label{fig:f10}}

	\caption{Sample images from the ARL Polarimetric Dataset.}
\end{figure}

\subsection{Network Architecture}
 For both datasets, the data corresponding to each modality goes into the corresponding encoder. The encoder projects the input modality into the feature space. Features are extracted from each modality independently, and are subsequently concatenated before going through the self-expressive layer. The input to the self-expressive represents the data modalities' projection into the latent space. The second component of the Volterra filter auto-encoder is the self-expressive layer. The goal of this layer is to enforce the self-representation property among the features extracted from each data modality, utilizing a fully connected layer which operates on the merged features. The last stage is the decoder which reconstructs input data from the self-expressive layers' output and has the same structure as the encoder.

 We implement the VNN-AE with Tensorflow and use the adaptive momentum based gradient descent method (ADAM) \cite{kingma2014adam} to minimize the loss function in Equation (2) with a learning rate of $10^{-3}$ for ARL dataset and $10^{-4}$ for EYB dataset. For DMSC, we use the same network structure described in \cite{abavisani2018deep}. In the following, we will elaborate on the network constructed for the VNN-AE for the two datasets: 
 
\subsubsection{ARL Dataset}
The ARL dataset consists of five data modalities, therefore, the auto-encoder has five encoder branches, one self expressive layers, and five decoder branches. Each encoder branch is composed of a single layered VNN (equivalent to a $2^{nd}$ order filter). The VNN consists of a multi-channel implementation inspired by the googlenet architecture on inception dataset \cite{szegedy2015going}. In our implementation, we use 3 channels with 1x1 filters and 2 channels with 3x3 filters. The decoder has the same structure as the encoder.
 
\subsubsection{EYB Dataset}
For EYB dataset, we use five data modalities, therefore, we have an encoder for each modality, one self expressive layers and five decoder branches. Each encoder is composed of a single layered VNN (equivalent to a $2^{nd}$ order filter). Each encoder consists of 7 channels with 1x1 filters, 7 channels with 3x3 filters, and 6 channels of 5x5 filters.
\subsection{Fusion Results}
We evaluate the performance of our proposed Volterra filter auto-encoder against the CNN based DMSC network.  The sparse solution $\mathbf{W}$ provides important information about the relations among data points, which may be used to split data into individual clusters residing in a common subspace. Observations from each object can be seen as data points spanning one subspace. Interpreting the subspace-based affinities based on $\mathbf{W}$, we proceed to carry out what amounts to modality fusion. For clustering by $\mathbf{W}$, we applied the same spectral clustering approach that we previously demonstrated in Section III.B. We used 75\% of the data to learn the cluster structure underlying the multi-modal data. We exploited three clustering metrics to evaluate the performance; the clustering accuracy rate (ACC), normalized mutual information (NMI) \cite{vinh2010information}, and Adjusted Rand Index (ARI) \cite{rand1971objective} metrics. The results are depicted in Tables I and II for EYB and ARL dataset respectively. From the results, it can be seen that VMSC-AE outperforms the DMSC network, all while reducing the number of parameters needed to carry out the clustering task. The reason behind this improvement is the fact that the VMSC-AE maintains a tractable structure that controls the non-linearities introduced in the system in contrast to the CNN network that can introduce undesirably infinite non-linearities.   
\begin{table}[!htb]
	\caption{Fusion Results for EYB Dataset}
	\begin{center}
		
		\begin{tabular}{|c|c|c|c|c|}
			\hline
			
			    & ACC & ARI & NMI & No. of Parameters  \\
			
			\hline 
			
			DMSC & 98.82\% &  98.08\%  & 98.81\%  & 2,367,400  \\ 
			\hline
			VFSC & 99.34\% & 98.63\%  & 99.15\% & 2,332,800 \\ 
			\hline

		\end{tabular}
		
		\label{tab6}
	\end{center}
\end{table}
\begin{table}[!htb]
	\caption{Fusion Results for ARL Dataset}
	\begin{center}
		
		\begin{tabular}{|c|c|c|c|c|}
			\hline
			
			& ACC & ARI & NMI & Number of Parameters  \\
			
			\hline 
			
			DMSC & 97.59\% & 97.53\%  & 99.42\%  & 4,667,720  \\ 
			\hline
			VFSC  & 99.95\% & 99.90\%  & 99.94\% & 4,666,650 \\ 
			\hline

		\end{tabular}
		
		\label{tab10}
	\end{center}
\end{table}
\subsection{Training with Less Data}
We proceed to evaluate the performance of the proposed Volterra filter auto-encoder with limited training data. A major challenge for any deep neural-network may be the availability of inadequately sufficient data to train the network. In the following, we will assess our proposed data fusion network versus the convolutional deep neural network, DMSC, in case of limited data availability during training. We train the auto-encoder structure using portions of the available data, i.e., 25\%, 40\%, 50\%, 60\%, and 75\%. The Results are depicted in Table III and IV for ARL and EYB dataset respectively. From the results, it is clear that fusing the data using VMSC-AE significantly boosts the clustering accuracy while using less parameters than DMSC. The VMSC-AE in case of ARL data uses just 520K parameters and 25\% of the data and still outperforms the DMSC model using 4.5M parameters and 75\% of the data for training. On the other hand, for EYB dataset, the performance of VMSC-AE with just 854K parameters and 40\% of data is comparable to that of the DMSC implementation with 2.4M parameters and 75\% data. Note that the number of parameters decreases with decrease in the amount of training data because the size of the self-expressive layer, $\mathbf{W}$ is directly dependent on the number of samples used for training the model. These results thus show that VMSC-AE is more robust and less sensitive to limited data availability during training.  
\begin{table*}[!htb]
	\caption{ARL Dataset: Training with Less Data}
	\begin{center}
		
		\begin{tabular}{|c|c|c|c|c|c|c|c|c|}
			\hline
			
			Dataset/Ratio & \multicolumn{3}{c|}{VFSC} & No. parameters & \multicolumn{3}{c|}{DMSC}& No. of parameters \\
			
			\hline 
			 & ACC & ARI & NMI &  & ACC & ARI& NMI&  \\
			
			\hline 
			
			ARL 25\% & 99.32\%& 98.19\% & 99.72\% & 519,450 & 93.33\% & 88.25\% & 97.86\% & 520,520  \\ 
			\hline
			ARL 40\% & 99.42\% & 99.49\% & 99.78\%  & 1,328,154 & 94\% & 91.98\% & 98.5\% & 1,329,224  \\ 
			\hline 
			ARL 50\% & 99.56\%  & 99.63\% & 99.86\% & 2,074,650 & 94.17\%  & 92.7\% & 98.6\% & 2,075,720 \\ 
			\hline 
			ARL 60\% & 99.9\% & 99.88\% & 99.94\%  & 2,987,034 & 95.69\%  & 93.87\% & 98.64\% & 2,988,104  \\ 
			\hline 
			ARL 75\% & 99.95\%  & 99.9\% & 99.95\% & 4,666,650 & 97.59\%  & 97.53\% & 99.42\% & 4,667,720 \\ 
			\hline 
			
		\end{tabular}
		
		\label{tab7}
	\end{center}
\end{table*}

The comparison between VFSC and DMSC performance for both ARL and EYB dataset are depicted in Figures 5 and 6 respectively.
\begin{figure}[!htb]
	\centering
	\includegraphics[width=9.5cm,height=7cm]{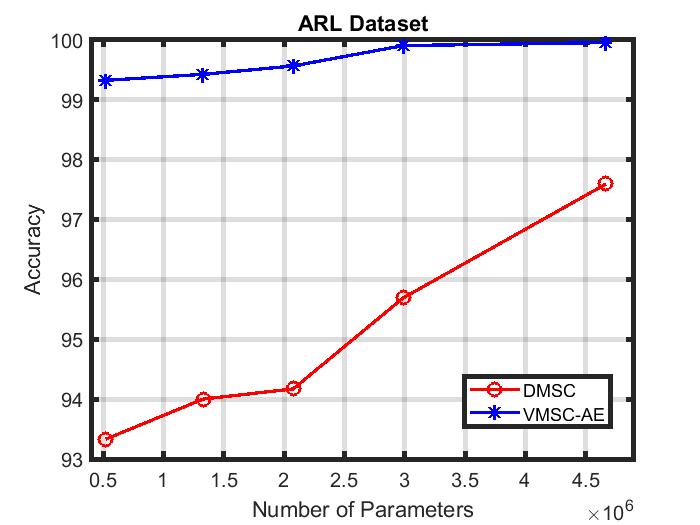}

	\caption{Comparison between VFSC and DMSC for ARL dataset.}.
\end{figure}

\begin{figure}[!htb]
	\centering
	\includegraphics[width=9.5cm,height=7cm]{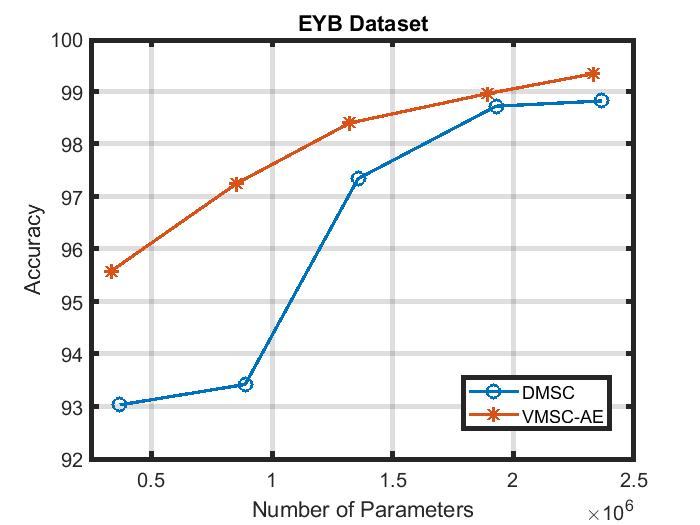}

	\caption{Comparison between VFSC and DMSC for EYB dataset.}.
\end{figure}

\begin{table*}[!htb]
	\caption{EYB Dataset: Training with Less Data}
	\begin{center}
		
	\begin{tabular}{|c|c|c|c|c|c|c|c|c|}
		\hline
		
		Dataset/Ratio & \multicolumn{3}{c|}{VFSC} & No. parameters & \multicolumn{3}{c|}{DMSC}& No. of parameters \\
		
		\hline 
		& ACC & ARI & NMI &  & ACC & ARI& NMI&  \\
		
		\hline 
			
			EYB 25\% & 95.58\% & 94.42\% & 96.8\%  & 333,784 & 93.33\% & 86.83\% & 92.34\% &  368,384  \\ 
			\hline
			EYB 40\% & 97.25\% & 96.81\% & 97.9\%  & 854,144 & 94\% & 93.21\% & 96.34\% &  888,744  \\ 
			\hline 
			EYB 50\% & 98.4\% & 96.9\% & 98.26\%  & 1,322,000 & 94.17\% & 95.05\% & 97.41\% &  1,356,600 \\ 
			\hline 
			EYB 60\% & 98.96\% & 97.99\% & 98.97\% & 1,893,824 & 95.69\% & 97.18\% & 98.47\% & 1,928,424 \\ 
			\hline 
			EYB 75\% & 99.34\% & 98.63\% & 99.15\% & 2,332,800 & 97.59\% &98.08\% & 98.81\% & 2,367,400\\ 
			\hline 
			
		\end{tabular}
		
		\label{tab8}
	\end{center}
\end{table*}
\subsection{Network Pruning by Random Removal of Edges}
 It is clear from Table 1 and 2 that the total number of parameters needed is dominated by the self-expressive layer parameters, which is a function of $O(N^2)$, where $N$ is the number of samples in the dataset. In this subsection, we  reduce the number of parameters by randomly removing a ratio of the edges in the self-expressive layer. In addition to training the network with less data, we reduce the number of edges needed to be trained by setting a fixed ratio of those edges to be equal zero and ignore them while training as if they do not exist. In Figures 7 and 8, we illustrate the error bar after removing 10\%, 30\%, 50\% and 70\% of the edges from the self-expressive layer for the ARL and EYB datasets respectively. The results have been averaged over 10 trials. From the results, we conclude that the VMSC-AE is more robust to the changes in the self-expressive layer connections as compared to DMSC. As we remove edges from the self-expressive layer, the performance degradation in case of VMSC-AE is more graceful. In addition, the Volterra Filter auto-encoder is less sensitive to training with less data as compared to DMSC network, as a result of lower number of parameters in the encoder and decoder which prevents over-fitting when lower number of samples are available.

\begin{figure*}[!htb]
	\begin{multicols}{2}
		\includegraphics[width=\linewidth]{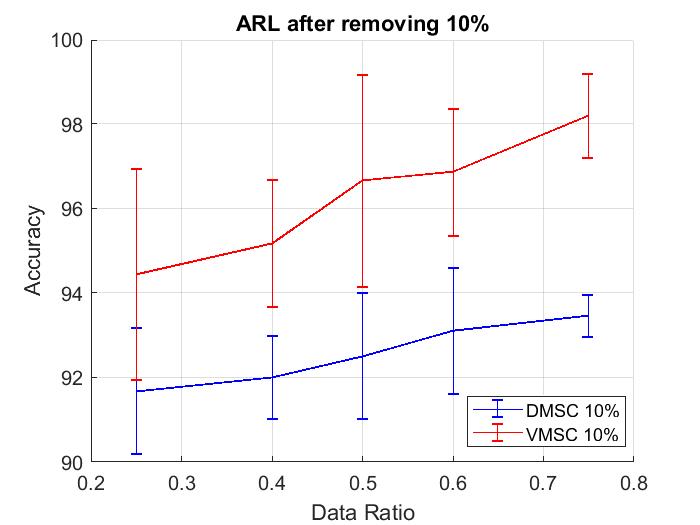}\par 
		\includegraphics[width=\linewidth]{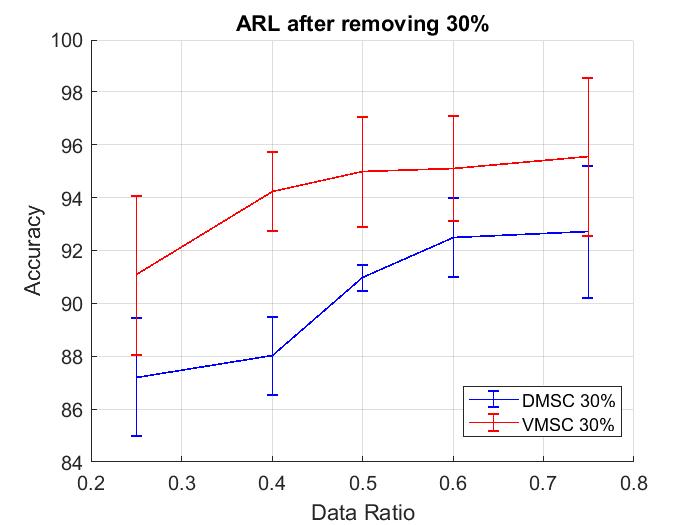}\par 
	\end{multicols}
	\begin{multicols}{2}
		\includegraphics[width=\linewidth]{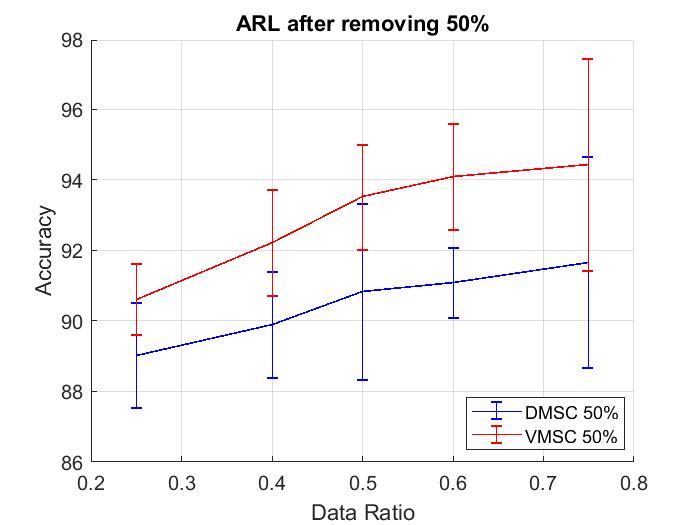}\par
		\includegraphics[width=\linewidth]{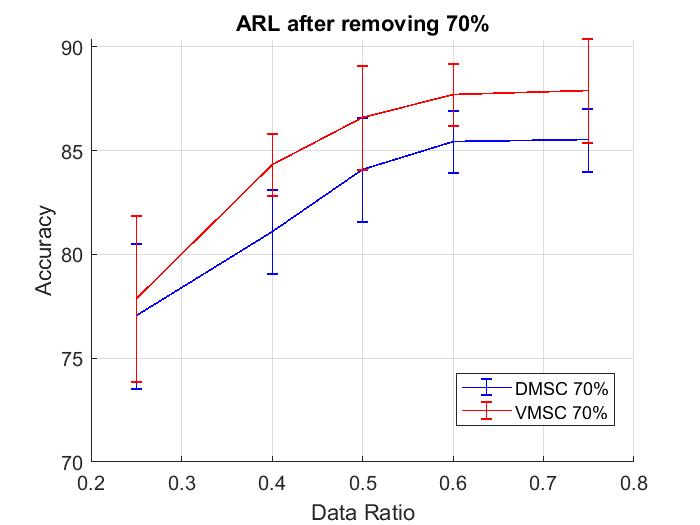}\par
	\end{multicols}
	\caption{Pruning the auto-encoder network while using different portions of the ARL data.}
\end{figure*}

\begin{figure*}[!htb]
	\begin{multicols}{2}
		\includegraphics[width=\linewidth]{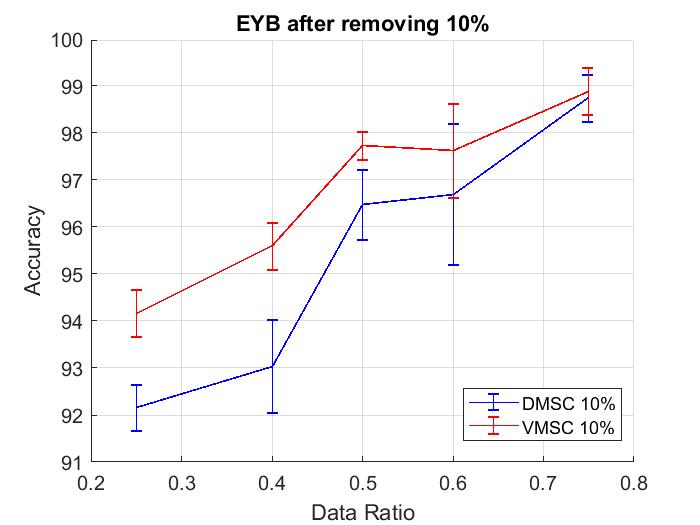}\par 
		\includegraphics[width=\linewidth]{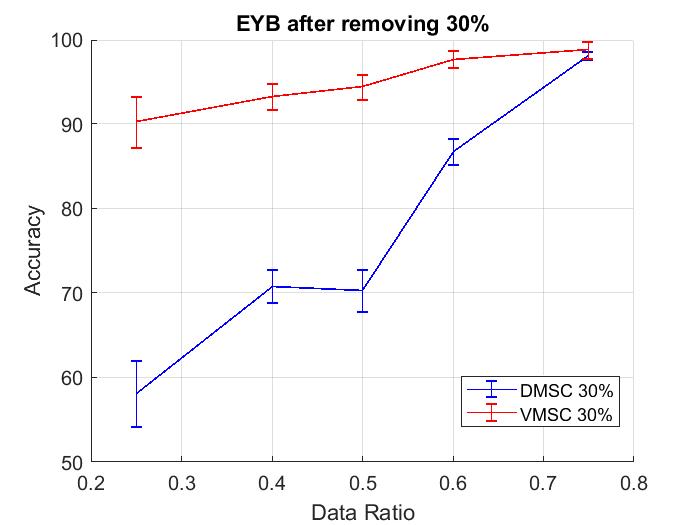}\par 
	\end{multicols}
	\begin{multicols}{2}
		\includegraphics[width=\linewidth]{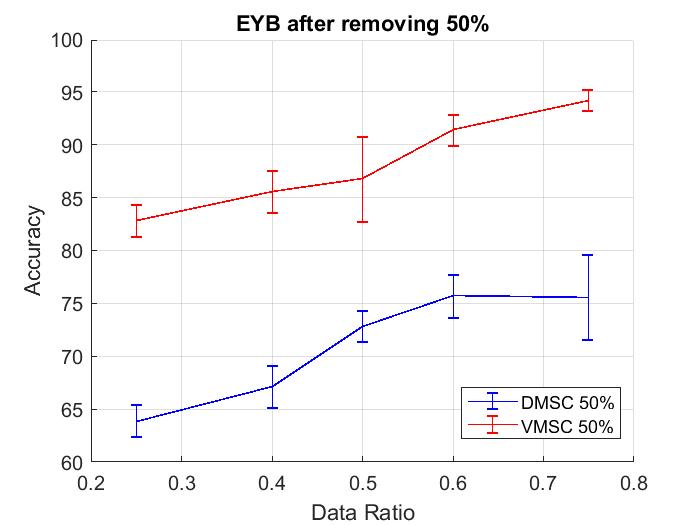}\par
		\includegraphics[width=\linewidth]{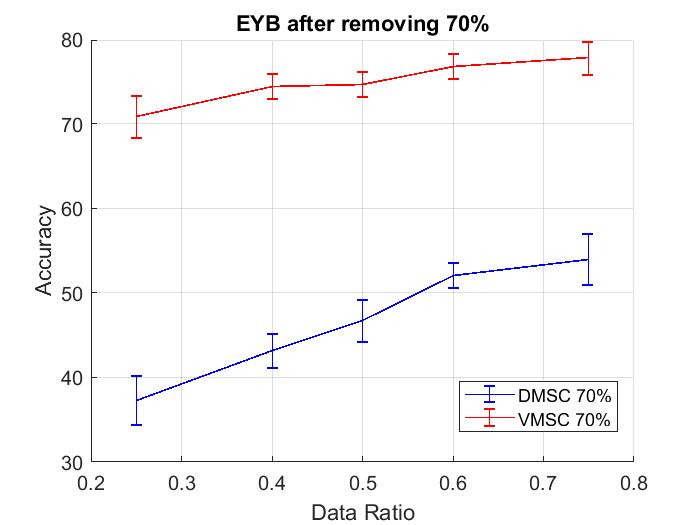}\par
	\end{multicols}
	\caption{Pruning the auto-encoder network while using different portions of the EYB data.}
\end{figure*}
\subsection{Network Pruning Using Cyclic Sparse Connected Layers}
In the following, we evaluate the CSC layers to further prune the auto-encoder network. As explained in section III, the CSC layers are described by the polynomial function by assigning each $p_i(x)$ to each support layer $i$. In our experiment, we assume that $C$=1 and given $N$, we compute the number of edges in each support layer accordingly using Eqn. (5). For the ARL dataset, we assume that $C = 1$ and $L = 2$ while for the EYB dataset, we take $L = 1$. We found out that this structure achieves the best performance while retaining a high compression rate. The experimental results are depicted in Table (4) and (5) for ARL and EYB datasets respectively. We utilized different data ratios for training the auto-encoder network and subsequently compare the results from the VMSC-AE network to the DMSC network. In addition, we list the number of parameters required for training the auto-encoder with the fully-connected self-expressive layer versus the number of parameters needed for training the auto-encoder utilizing the CSC layers. The results are also illustrated in Figures 9 and 10 for ARL Dataset and EYB Dataset respectively.

\begin{figure}[!htb]
	\centering
	\includegraphics[width=9.5cm,height=7cm]{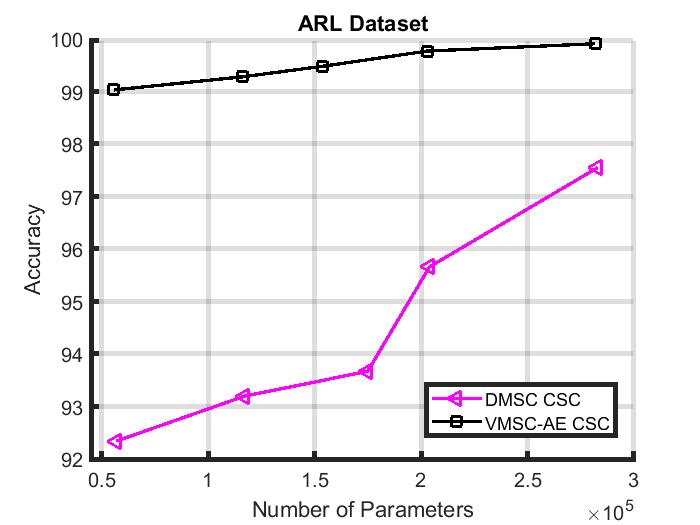}

	\caption{Performance VS. Number of parameters of VFSC CSC, and DMSC CSC for ARL dataset.}.
\end{figure}

\begin{figure}[!htb]
	\centering
	\includegraphics[width=9.5cm,height=7cm]{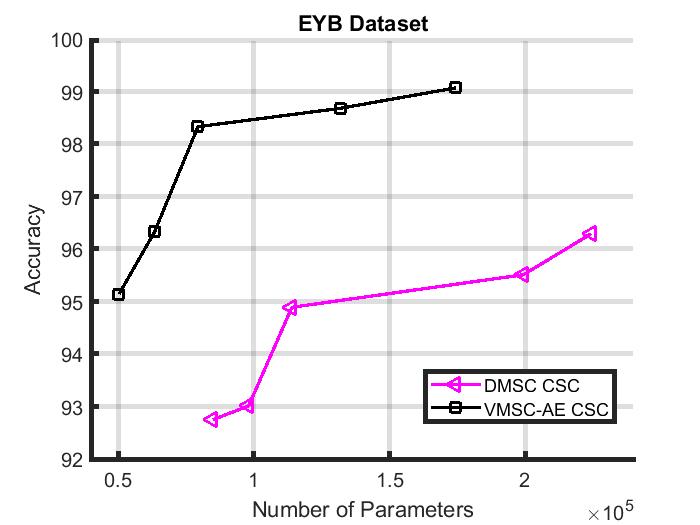}

	\caption{Performance VS. NUmber of parameters of VFSC CSC, and DMSC CSC for EYB dataset.}.
\end{figure}

\begin{table*}[!htb]
	\caption{ARL: Impact of CSC layers when training with less data.}
\begin{center}
	
	\begin{tabular}{|c|c|c|c|c|c|c|c|c|}
		\hline
		
		Dataset$|$Ration & \multicolumn{3}{c|}{Fully Connected} & No. parameters & \multicolumn{3}{c|}{CSC Layers} & No. of parameters \\
		
		\hline 
		 & ACC & ARI & NMI &  & ACC & ARI & NMI & \\
		
		\hline 
		
		ARL DMSC 25\% & 93.33\% & 88.25\% & 97.86\% & 520,520 & 92.34\% & 85.1\% & 97.54\%  & 56,840  \\ 
		\hline
		ARL VFSC 25\% & 99.32\% & 98.19\% & 99.72\% & 519,450 & 99.03\% & 97.19\% & 99.16\% & 55,770  \\ 
		\hline 
		ARL DMSC 40\% & 94\% & 91.98\% & 98.5\% & 1,329,224 & 93.2\% & 90.24\% & 98.13\%  & 117,320 \\ 
		\hline 
		ARL VFSC 40\% & 99.42\% & 99.49\% & 99.78\%  & 1,328,154 & 98.75\%  & 97.67\% & 99.27\% & 116,250 \\ 
		\hline 
		ARL DMSC 50\% & 94.17\%  & 92.7\% & 98.6\% & 2,075,720 & 93.67\%  & 92.05\% & 98.22\% & 174,920 \\ 
		\hline 
		ARL VFSC 50\% & 99.56\%  & 99.63\% & 99.86\% & 2,074,650 & 99.48\%  & 99.01\% & 99.55\% & 153,690 \\ 
		\hline
		ARL DMSC 60\% & 95.69\%  & 93.87\% & 98.64\% & 2,988,104 & 95.66\%  & 92.89\% & 98.4\% & 203,960 \\ 
		\hline 
		ARL VFSC 60\% & 99.9\%  & 99.88\% & 99.94\% & 2,987,034 & 99.77\%  & 99.53\% & 99.71\% & 202,890 \\ 
		\hline 
		ARL DMSC 75\% & 97.59\%  & 97.53\% & 99.42\% & 4,667,720 & 97.55\%  & 97.81\% & 99.47\% & 282,920 \\ 
		\hline 
		ARL VFSC 75\% & 99.95\%  & 99.9\% & 99.95\% & 4,666,650 & 99.91\%  & 99.8\% & 99.93\% & 281,850 \\ 
		\hline

	\end{tabular}
	
	\label{tab11}
\end{center}
\end{table*}

\begin{table*}[!htb]
	\caption{EYB Dataset: Impact of CSC layers when training with less data.}
	\begin{center}
		
		\begin{tabular}{|c|c|c|c|c|c|c|c|c|}
			\hline
			
			Dataset$|$Ration & \multicolumn{3}{c|}{Fully Connected} & No. parameters & \multicolumn{3}{c|}{CSC Layers} & No. of parameters \\
			
			\hline 
			& ACC & ARI & NMI &  & ACC & ARI & NMI & \\
			
			\hline 
			
			EYB DMSC 25\% & 93.03\% & 86.83\% & 92.34\% & 368,384 & 92.75\%  & 82.25\% & 91.74\% & 84,900  \\ 
			\hline
			EYB VFSC 25\% & 95.58\%  & 94.42\% & 96.8\% & 333,784 & 95.14\%  & 91.85\% & 95.92\% & 50,300  \\ 
			\hline 
			EYB DMSC 40\% & 93.42\%  & 93.21\% & 96.34\% & 888,744 & 93.01\%  & 86.43\% & 93.41\% & 98,040 \\ 
			\hline 
			EYB VFSC 40\% & 97.25\%  & 96.81\% & 97.9\% & 854,144 & 96.33\%  & 96.12\% & 97.43\% & 63,440 \\ 
			\hline 
			EYB DMSC 50\% & 97.34\%  & 95.05\% & 97.41\% & 1,356,600 & 94.89\%  & 88.43\% & 94.78\% & 114,000\\ 
			\hline  
			EYB VFSC 50\% & 98.4\%  & 96.9\% & 98.26\% & 1,322,000 & 98.33\%  & 96.71\% & 98.22\% & 79,400  \\ 
			\hline
			EYB DMSC 60\% & 98.72\%  & 97.18\% & 98.47\% & 1,928,424 & 95.51\%  & 91.59\% & 95.72\% & 199,272  \\ 
			\hline 
			EYB VFSC 60\% & 98.96\%  & 97.99\% & 98.97\% & 1,893,824 & 98.68\%  & 97.42\% & 98.5\% & 131,840 \\ 
			\hline 
			EYB DMSC 75\% & 98.82\%  & 98.08\% & 98.81\% & 2,367,400 & 96.29\%  & 96.93\% & 98.32\% & 224,200 \\ 
			\hline 
			EYB VFSC 75\% &  99.34\% & 98.63\% & 99.15\% & 2,332,800 & 99.07\%  & 98.09\% & 98.81\% & 174,400\\ 
			\hline 
			
		\end{tabular}
		
		\label{tab9}
	\end{center}
\end{table*}

From the results, we can conclude that VMSC-AE significantly outperform the CNN-based autoencoder, DMSC, after pruning the self-expressive layer using CSC layer structure. VMSC-AE could achieve a very competitive performance with almost 1\%-2\% accuracy loss as compared to fully connected network. Moreover, the model complexity has decreased by slightly over $>90$\%.
\section{Conclusion}
In this paper, we presented an efficient Volterra Neural Network auto-encoder for multi-modal data fusion. The introduced framework extracted the underlying embedding of each data modality under the assumption of data self-representation. Experimental results show a significant improvement for Volterra Neural Network over the Convolutional Neural Network auto-encoder. In addition, we evaluated multiple approaches to further prune the network structure and reduce the model complexity of the multi-modal subspace clustering auto encoder method. The experimental results showed that our proposed approach provides better sample complexity over CNN-based auto-encoder and demonstrates a robust classification performance.


%

\appendices

\bibliographystyle{IEEEtran}
\bibliography{deep2}
\begin{IEEEbiographynophoto}{Sally Ghanem}
received the B.Sc. degree in electrical engineering from Alexandria University, in 2013, and the M.Sc. degree in electrical and computer engineering from North Carolina State University, Raleigh, NC, USA in 2016, where she is currently pursuing the Ph.D. degree. During the PhD program, she has spent time at Oak Ridge National Laboratory working on multimodal data. Her research interests include the areas of computer vision, digital signal processing, image processing and machine learning.
\end{IEEEbiographynophoto}

\begin{IEEEbiographynophoto}{Siddharth Roheda}
	received the B.Tech. degree in Electronics and Communication from Nirma University, India, in 2015, and the M.Sc. and Ph.D. degrees from the Department of Electrical and Computer Engineering, North Carolina State University, Raleigh, NC, USA, in 2020. He is currently a Post-Doctoral Researcher at North Carolina State University working with the Vision, Information, and Statistical Signal Theories and Applications Group. His current research interests include Robust Information Fusion, Machine Learning, Deep Learning, Computer Vision, and Signal Processing. More specifically, he is interested in Supervised Learning, developing novel deep learning architectures and robust algorithms for exploiting multi-modal information for target detection and recognition.
\end{IEEEbiographynophoto}

\begin{IEEEbiographynophoto}{Hamid Krim}
received the B.Sc. and M.Sc. and Ph.D. in ECE. He was a Member of Technical Staff at AT\&T Bell Labs, where he has conducted research and development in the areas of telephony and digital communication systems/subsystems. Following an NSF Postdoctoral Fellowship at Foreign Centers of Excellence, LSS/University of Orsay, Paris, France, he joined the Laboratory for Information and Decision Systems, MIT, Cambridge, MA, USA, as a Research Scientist and where he performed/supervised research. He is currently a Professor of electrical engineering in the Department of Electrical and Computer Engineering, North Carolina State University, NC, leading the Vision, Information, and Statistical Signal Theories and Applications Group. His research interests include statistical signal and image analysis and mathematical modeling with a keen emphasis on applied problems in classification and recognition using geometric and topological tools. He has served on the SP society Editorial Board and on TCs, and is the SP Distinguished Lecturer for 2015-2016.
\end{IEEEbiographynophoto}

%





\end{document}